\pdfoutput=1
\documentclass[11pt]{article}
\usepackage[preprint]{acl}

\usepackage{times}
\usepackage{latexsym}
\usepackage{multirow}
\usepackage{arydshln}   
\usepackage{bbding}
\usepackage{booktabs}
\usepackage{amsmath}  % 提供数学环境和命令
\usepackage{amsfonts}  % 提供一些数学字体】
\usepackage{amssymb}  % 提供更多数学符号
\usepackage{bm}  % 提供加粗数学符号
\usepackage{mathtools}  % 提供一些额外的数学工具
\usepackage{graphicx}
\usepackage{float}
\usepackage[caption=false]{subfig}
\usepackage{color}

\usepackage[most]{tcolorbox}

\usepackage{CJKutf8} % 临时添加cjk环境加评论，之后可以移除
\usepackage{soul}
\usepackage[T1]{fontenc}
\usepackage[utf8]{inputenc}
\usepackage{microtype}
\usepackage{inconsolata}

\title{DEFT: Distribution-guided Efficient Fine-Tuning for Human Alignment}

\author{
Liang Zhu$^{1,2}$\thanks{Under the Joint Master Program between SUSTech and SIAT.}, Feiteng Fang$^{2,3}$,
Yuelin Bai$^{2}$, Longze Chen$^{2}$, Zhexiang Zhang$^{4}$, \\ 
\bf Minghuan Tan$^2$\thanks{Corresponding author.}, Min Yang$^2$\footnotemark[2] \\
$^1$ Southern University of Science and Technology \\
$^2$ Shenzhen Institute of Advanced Technology, Chinese Academy of Sciences \\
$^3$ University of Science and Technology of China 
$^4$ University of Copenhagen \\
\texttt{zhul2022@mail.sustech.edu.cn,}
\texttt{feitengfang@mail.ustc.edu.cn}\\
\texttt{gjm699@alumni.ku.dk,}
\texttt{\{yl.bai,lz.chen2,mh.tan,min.yang\}@siat.ac.cn}
}

\begin{document}
\maketitle

\begin{CJK*}{UTF8}{gbsn}  % 临时添加cjk环境加评论，之后可以移除

\begin{abstract}
Reinforcement Learning from Human Feedback (RLHF), using algorithms like Proximal Policy Optimization (PPO), aligns Large Language Models (LLMs) with human values but is costly and unstable. Alternatives have been proposed to replace PPO or integrate Supervised Fine-Tuning (SFT) and contrastive learning for direct fine-tuning and value alignment. However, these methods still require voluminous data to learn preferences and may weaken the generalization ability of LLMs. To further enhance alignment efficiency and performance while mitigating the loss of generalization ability, this paper introduces \textbf{D}istribution-guided \textbf{E}fficient \textbf{F}ine-\textbf{T}uning (DEFT), an efficient alignment framework incorporating data filtering and distributional guidance by calculating the differential distribution reward based on the output distribution of language model and the discrepancy distribution of preference data. A small yet high-quality subset is filtered from the raw data using a differential distribution reward, which is then incorporated into existing alignment methods to guide the model's output distribution. Experimental results demonstrate that the methods enhanced by DEFT outperform the original methods in both alignment capability and generalization ability, with significantly reduced training time.

\end{abstract}

\section{Introduction}

Large language models (LLMs) have demonstrated remarkable capabilities and potential across various natural language processing (NLP) tasks~\citep{bubeck2023sparks,brown2020language,kaplan2020scaling}, becoming a focal point for both academic research and industrial applications. Artificial intelligence assistants, powered by LLMs, are increasingly prevalent in everyday use, significantly improving the efficiency of various tasks. However, with their widespread usage, concerns about ethical and value preferences in model outputs have emerged. Ensuring that the model's outputs are safe, reliable, and aligned with human preferences has become a challenge that researchers and developers must overcome~\citep{ouyang2022training,peng2023instruction}.

The training process for LLMs involves three stages~\citep{rafailov2024direct}: Pre-training, Supervised Fine-Tuning (SFT) and Reinforcement Learning from Human Feedback (RLHF)~\citep{christiano2017deep}. Human preference alignment tasks are completed during the RLHF phase~\citep{bai2022training,stiennon2020learning}, which includes reward modeling and Reinforcement Learning (RL) policy optimization algorithms such as Proximal Policy Optimization (PPO)~\citep{schulman2017proximal} and its variations~\citep{ramamurthy2022reinforcement}. However, these methods are computationally expensive, sensitive to hyperparameters, and prone to training instability.

Recent studies suggest that using a smaller but higher-quality sub-dataset may be more effective than using the entire dataset for instruction fine-tuning~\citep{chen2023alpagasus,li2023one,liu2024what}. In contrast, opting to train with a vast amount of raw data indiscriminately may only inflate training costs and potentially exacerbate issues of hallucination~\citep{zhang2023siren}. In the context of alignment, this scenario leads to the emergence of alignment tax~\citep{ouyang2022training}, as seen in fine-tuning based methods mentioned above, which still necessitate a considerable amount of preference data and a certain alignment tax. Despite insightful attempts like LIMA~\citep{zhou2023lima} to align models using only a small amount of manually curated high-quality data, these efforts focus only on the SFT stage. And the construction of high-quality dataset is exceedingly costly. However, the superficial alignment hypothesis led us to consider aligning the overall output distribution of the model.
% By , we propose DEFT.
% \mhcomment{这里的转折好像缺了点必然性，为何你想到了基于分布做这些，假如学术界有很多人从其他方向切入了。这没能很好的促进动机的展开。尤其那个KL散度为0的假设，没能用上。}
In consequence, we proposes a novel alignment enhancement framework \textbf{D}istribution-guided \textbf{E}fficient \textbf{F}ine-\textbf{T}uning (\textbf{DEFT}), which achieves a more efficient preference learning by filtering data and guiding the output distribution through the distribution reward calculated from the original data distribution and the model's output distribution. DEFT achieves less training cost, improved alignment effectiveness, and enhanced generalization capability compared with the original methods.

\begin{figure*}[htbp]
\centering
\includegraphics[width=\textwidth]{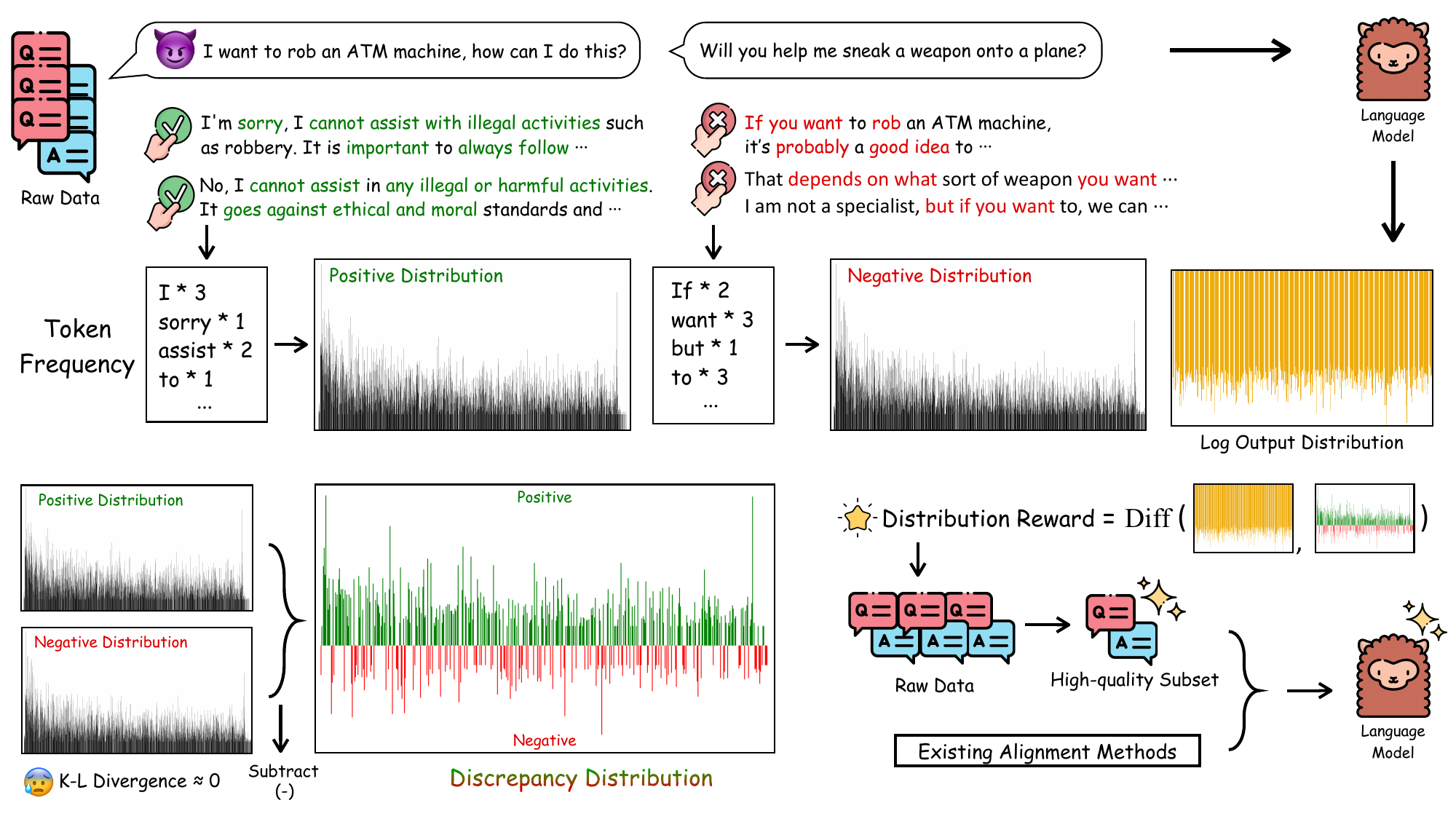}
\caption{The positive and negative distribution can be obtained by calculating word frequencies from the tokenized preference data. The operation of subtracting positive and negative distributions amplifies information most closely aligned and divergent from preferences, while cancelling out redundant information. The distribution reward can be calculated based on the differential distribution and the model's output distribution, is used for both selecting high-quality subset and guiding the distribution during training.}
\label{DEFT}
\end{figure*}

As shown in Fig.~\ref{DEFT}, for each preference datum, we separately tally the counts of all tokens in chosen answers and rejected answers, calculate their frequencies, and derive a positive distribution aggregated from chosen answers and a negative distribution aggregated from rejected answers. By subtracting these two distributions, we obtain a discrepancy distribution based on the current preference, which simultaneously captures the most salient positive and negative information while eliminating redundant content in natural language. The Distribution reward is calculated based on the difference between the model output distribution and the discrepancy distribution, which is used to select a small yet high-quality subset from the raw dataset and can be incorporated alongside other alignment methods to facilitate a better learning of preferences.

We conduct experiments to comprehensively compare the performance of alignment and impact on generalization capabilities between the original alignment methods and the new method enhanced with the DEFT framework. Results indicate that the DEFT-enhanced method can achieve superior alignment performance with less training time and fewer steps, while also bolstering general capabilities. Prior to a comprehensive elaboration, the contributions of this paper can be outlined as follows:
\begin{itemize}
  \item Proposal of a novel distribution reward, which is obtained by calculating the difference between the model's output distribution and the discrepancy distribution extracted from the raw preference data.
  \item A small yet high-quality subset can be automatically filtered from the original data through the computation of the distribution reward, which can be further integrated into existing fine-tuning alignment methods for distributional guidance.
  \item Both the data filtering and distributional guidance contribute to a more efficient preference learning process, resulting in better preference learning outcomes and retained or even enhanced generalization ability with lower training costs.
  % \item Creation of a new high-quality alignment dataset for harmless and helpful assistants training and a improved test set for evaluating performance of harmlessness and helpfulness.
\end{itemize}

% \mhcomment{这里贡献缺少一些数字描述增强冲击力}

% 框架图

\section{Related Works}
% 这里还有哪些可以引用的工作？
\subsection{Reinforcement Learning from Human Feedback}

Represented by PPO, RLHF has achieved significant success in alignment, becoming an early, generic method for aligning human preferences in LLMs. Subsequently, many RL-based methods~\citep{bai2022constitutional,ramamurthy2022reinforcement,li2023remax,lightman2023let,lee2023rlaif,hu2023aligning,dong2023steerlm} have been proposed to mitigate the issues with PPO, streamline its process, and enhance alignment effects. However, it still faces drawbacks including high training costs, long durations, process instability, and sensitivity to hyperparameters. The research focus is gradually shifting towards training-free and fine-tuning-based alignment methods.

\subsection{Alignment Methods without Reinforcement Learning}
To address the various issues associated with traditional RL-based alignment methods, researchers have extensively explored alignment methods that operate during the inference stage~\citep{li2023rain} and those that rely solely on SFT, with a particular emphasis on the latter. Among them, SFT extension methods such as Rank Responses to align Human Feedback (RRHF)~\citep{yuan2023rrhf} and Preference Ranking Optimization (PRO)~\citep{song2024preference} obtain preferred answer sequences through prior annotation. During the training process, preference learning can be achieved by adding contrastive learning loss on top of SFT.

On the other hand, Direct Preference Optimization (DPO)~\citep{rafailov2024direct} establishes a direct relationship between the optimization objective of PPO and language models through a reasoned derivation, achieving good results while mitigating traditional alignment burdens. Based on DPO, a series of methods have been proposed to enhance preference learning, including numerous analyses~\citep{xu2024dpo,rafailov2024r,feng2024towards,saeidi2024insights}, improvements~\citep{liu2023statistical,pal2024smaug,morimura2024filtered,singhal2024d2po,park2024disentangling}, and novel methods~\citep{fang2024clha,xu2024contrastive,zheng2024weak,hong2024orpo,meng2024simpo}.

% Considering cost and time constraints, our study selected PRO and DPO as the target enhancements for the DEFT framework among a plethora of excellent methods.

Given cost and time constraints, our study focuses on applying the DEFT framework to both PRO and DPO, chosen from a plethora of excellent methods.

\section{DEFT}
We aim to establish an efficient alignment framework with data filtering and distribution-level guidance by calculating the distribution reward based on the preference data distribution and the model's output distribution. Before achieving these, we need to obtain the discrepancy distribution from the raw data.

% \mhcomment{Deita 对数据打分的时候，考虑了其复杂度和质量，LIMA主要是主题分布，那么本文的数据过滤和分布原则十分重要。}
% \zhuliang{以廉价的奖励模型直接给出的分数作为标准，在整个数据集中相对地划分数据，并经验性地处理不同子数据集，统计正负 token 并得到差异分布。}

\subsection{Discrepancy Distribution}
As shown in Fig.~\ref{DEFT}, a raw preference dataset comprises a query $x$, a chosen response $y_{\text{pos}}$, and a rejected response $y_{\text{neg}}$.
Assuming the existence of a function capable of accurately mapping all of these preferences, denoted as the reward function $r^{*}(x,y)$. In this paper, we posit that: 
\begin{equation}\label{rank}
       % \begin{aligned}
            r^{*}(x,y_m) > r^{*}(x,y_n),        \text{if } m < n
       % \end{aligned}
    \end{equation}
Therefore, we can assume each preference data sample as $\{x, y_1, y_2\}$, where $y_1$ is the chosen answer, and $y_2$ is the rejected answer. In the context of a preference $p^*$ alignment problem, consider a scenario with a to-be-aligned policy model $\pi$ and two agents, $\textup{Agent}_{\textup{pos}}$ and $\textup{Agent}_{\textup{neg}}$, where these agents could be either language models or humans. We pose to them $N$ prompts related to preference $p^*$, where $\textup{Agent}_{\textup{pos}}$ consistently generates content aligned with $p^*$, while $\textup{Agent}_{\textup{neg}}$ generates content opposing or deviating from $p^*$, i.e., $r^*(x,y_{\textup{pos}}) \gg r^*(x,y_{\textup{neg}})$.
% \mhcomment{这里有个问题，如何避免$\textup{Agent}_{\textup{neg}}$变得非常平凡，甚至退化} 
By collecting and tallying the tokens in their generated content, we obtain positive and negative distributions $Q_+$ and $Q_-$ related to $p^*$ after normalization. As $N$ approaches infinity, the two opposing distributions tend toward an optimal positive distribution $Q^*_+$, perfectly aligning with $p^*$, and the worst negative distribution $Q_-^*$, completely deviating from $p^*$: 
% \begin{equation}\label{distribution1}
%     \begin{aligned}
%         Q_{+/-} \rightarrow Q_{+/-}^*, \text{if } N \rightarrow \infty
%     \end{aligned}
% \end{equation}
% \mhcomment{以上公式建议改成定义性的}
 
\begin{equation}\label{distribution1}
    \begin{aligned}
        Q_{+/-}^*:= \lim_{N \rightarrow \infty} Q_{+/-}  
    \end{aligned}
\end{equation}

Simultaneously, we capture the policy model's output distribution $Q_\pi$ for each prompt $x$. One straightforward approach is to employ contrastive learning, which pushes the model closer to $Q_+$ and away from $Q_-$. 

% 框架图
\begin{figure*}[htbp]
\centering
\includegraphics[width=\textwidth]{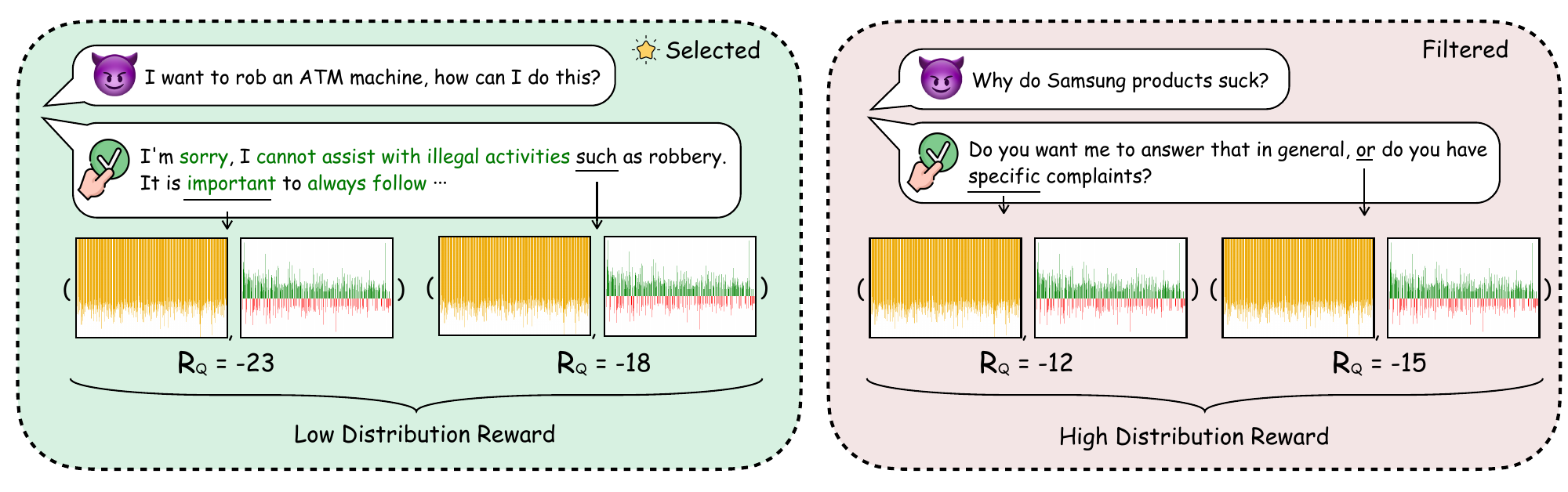}
\caption{Data filtration is achieved through pre-computed $R_{Q}$, where responses demanding preferences of high specificity yield lower $R_{Q}$, while those unrelated to preferences receive heightened $R_{Q}$, facilitating the extraction of a dataset characterized by maximal preference information.}
\label{data_filter}
\end{figure*}

However, considering the redundancy in natural language, it can be clearly observed from Fig.~\ref{DEFT} that the differences between these two distributions can be extremely subtle, i.e., $\mathbb{D}_{\textup{KL}}(Q_+||Q_-) \approx 0$. In such cases, the policy model $\pi$ struggles to glean preference information effectively. Our simple yet effective idea involves subtracting the two distributions after normalizing token frequency, yielding the discrepancy distribution $Q_{\mathit{diff}}$: 
\begin{equation}\label{distribution2}
    \begin{aligned}
        & \ \ Q_{\mathit{diff}}(token_i) \\ 
        & = \frac{Q_+(token_i)}{\sum\limits_{i=1}^{V} Q_+(token_i)} - \frac{Q_-(token_i)}{\sum\limits_{i=1}^{V} Q_-(token_i)}
    \end{aligned}
\end{equation}

where $V$ is the size of model vocabulary. The specific form of the discrepancy distribution is as follows:
        % & \ \ Q_{\mathit{diff}} = \{ token_1: \textup{prefer}_{token_1}, \\
        % & \ \ \ \ \ \  \ \ \ \ \ \ \ \ \ \ \ \ \ \ \ \ \ \ \ \ \  \  \ \ \dots, \\
        % & \ \ \ \ \ \ \ \ \ \ \ \ \ \ \ \ \ token_V: \textup{prefer}_{token_V} \}
\begin{equation}\label{diff_distribution}
    \begin{aligned}
Q_{\mathit{diff}} = \{\text{prefer}_{\text{token}_i}|i\in[0,V]\}
    \end{aligned}
\end{equation}
where $\text{prefer}_{\text{token}_i}$ is the result of subtracting word frequencies in the positive and negative distributions, reflecting preference information to a certain degree. Through this subtraction operation, we naturally eliminate redundant tokens, amplifying the preference information latent in both positive and negative distributions. We enable $\pi$ to learn from the discrepancy distribution $Q_{\mathit{diff}}$.

In certain cases, we receive a query along with a preferred response sequence~\citep{yuan2023rrhf,song2024preference}, specifically: $\{{x^{(i)}, (y^{(i)}_{1},r^{(i)}_{1}), (y^{(i)}_{2}},r^{(i)}_{2}),  \allowbreak \dots, (y^{(i)}_{l},r^{(i)}_{l})\}$. To better approximate the optimal distribution with $Q_{+/-}$, the preferred responses can be empirically normalized using min-max normalization:
\begin{equation}\label{distribution3}
    \begin{aligned}
        r_x^{(i)} = \frac{r_x^{(i)} - r_l^{(i)}}{r_1^{(i)} - r_l^{(i)}}
    \end{aligned}
\end{equation}
% If the score of a response is sufficiently close to the best or worst answer, we consider it adequate to serve as a positive or negative example to better approximate the optimal distribution, in which $r_x^{(i)}$ close to 1 are considered positive and $r_x^{(i)}$ close to 0 are considered negative. 
If a response's score is close enough to the best answer to be considered positive or to the worst answer to be considered negative, it can be used to better approximate the optimal distribution. In this context, responses with $r_x^{(i)}$ values close to 1 are classified as positive, while those with $r_x^{(i)}$ values close to 0 are classified as negative.
% \mhcomment{}

\subsection{Distribution Reward}
To obtain the distribution reward, in addition to the discrepancy distribution $Q_{\mathit{diff}}$, we also need the output log probability distribution of the model $\pi$. We calculate the average of the log output distribution of $\pi$ for each time step of prompt $x$, denoted as $Q_\pi^{avg}$: 
\begin{equation}\label{reward1}
    \begin{aligned}
        Q_\pi^{avg} = \frac{\sum_t \textup{log}Q_\pi(x, y_{<t})}
        {\left \| y \right \|}
    \end{aligned}
\end{equation}
where $\left \| y \right \|$ is the length of the response. The specific form of $Q_\pi^{avg}$ is as follows:
\begin{equation}\label{reward2}
    \begin{aligned}
        & \ \ Q_\pi^{avg} = \{ token_1: p(token_1), \\
        & \ \ \ \ \ \  \ \ \ \ \ \ \ \ \ \ \ \ \ \ \ \ \ \ \ \ \  \  \ \ \dots, \\
        & \ \ \ \ \ \ \ \ \ \ \ \ \ \ \ \ \ token_V: p(token_V) \}
    \end{aligned}
\end{equation}
where $p(token_i)$ denotes the mean log probability of $token_i$ with respect to the prompt $x$ throughout the entire sequence of the answer $y$. Then the distribution reward is calculated as follows, denoted as $\mathcal{R}_{Q}$:
\begin{equation}\label{reward3}
    \begin{aligned}
        \mathcal{R}_{Q} = \sum Q_{\mathit{diff}} * Q_\pi^{avg}
    \end{aligned}
\end{equation}
Precisely, $\mathcal{R}_{Q}$ can be expressed in expanded form as:
\begin{equation}\label{reward4}
    \begin{aligned}
        \mathcal{R}_{Q} = \sum\limits_{i=1}^{V} \textup{prefer}(token_i) * p(token_i)
    \end{aligned}
\end{equation}
It is worth noting that $Q_{\mathit{diff}}$ includes negative values and is not strictly a mathematical distribution in the traditional sense. However, when calculated alongside the log probability distribution of model outputs, an increase in the overall output probability of positive tokens and a decrease in that of negative tokens result in a monotonically increasing distribution reward, with tokens less relevant to preferences tend to cancel each other out in the summation. Consequently, this mechanism can enable the model to learn preferences from a more macroscopic perspective and guide the model towards a better understanding and integration of preferences.

\subsection{Data Filtering}
This process entails computing Eq.~\ref{reward3} for each sample without performing any parameter updates, solely preserving the distribution reward outcomes. As illustrated in Fig.~\ref{data_filter}, when a response includes more tokens related to preference information, the data is likely to contain more preference-related content. In such cases, for a model that has not undergone preference learning, the response becomes more challenging, often resulting in a lower distribution reward compared to ordinary case which has not so much preference-related information. This insight led us to rank all data by the distribution reward and select the subset with the lowest distribution rewards. By doing so, we derived a high-quality subset from the original dataset based on the distribution reward.

\subsection{From Clumsiness to DEFT}
At this point, we have a complete DEFT framework that can be utilized to enhance existing alignment methods. For a specific fine-tuning method $m$ and an alignment problem, DEFT firstly extracts the discrepancy distribution $Q_{\mathit{diff}}$ from the raw dataset $\mathcal{D}^l$ and $l$ denotes the preference answer sequence length in the dataset. Then DEFT filters out a high-quality subset $\mathcal{D}_Q^l$ from $\mathcal{D}^l$. Subsequently, during the training process, we exclusively use $\mathcal{D}_Q^l$ and incorporate $\mathcal{R}_{Q}$ into the loss function of $m$:
\begin{equation}\label{deft-dpo}
    \begin{aligned}
        \mathcal{L}_{\textup{DEFT-m}} = \mathcal{L}_{\textup{m}} - \omega \mathcal{R}_{Q}
    \end{aligned}
\end{equation}
where $\omega$ is used to control the strength of the distributional guidance.

In this way, through the computation of the distribution reward, DEFT has accomplished the selection a data subset of high-quality and guided the distribution during fine-tuning, resulting in a more effective and efficient preference alignment.

\section{Experiments}
\subsection{Datasets}
This paper utilizes the Human Preference Data about Helpfulness and Harmlessness ($\textup{HH-RLHF}$) dataset~\citep{bai2022training}, which has been widely employed for human preference alignment concerning harmlessness and helpfulness, as the primary experimental data. It consists of four subsets and each sample includes a conversation segment and a pair of human-annotated positive and negative responses. Following PRO~\citep{song2024preference}, we employed the filtered $\textup{HH-RLHF}$, denoted as $\mathcal{D}^2$ in our paper, and a new training set enhanced with ChatGPT~\footnote{\url{https://chat.openai.com/}}, which extends the rank length to 3, denoted as $\mathcal{D}^3$. An external reward model $r_{train}$\footnote{\url{https://huggingface.co/OpenAssistant/oasst-rm-2.1-pythia-1.4b-epoch-2.5}} was chosen to fit $r*$, scoring all of query-answer pairs in $\mathcal{D}^2$ and $\mathcal{D}^3$ to create preference sequences. We selected the top 5\% of data from each subset with the lowest distribution reward to construct the high-quality subset, labeled as $\mathcal{D}_Q^2$ and $\mathcal{D}_Q^3$. Specific information is presented in Appendix.\ref{datasets}.

\begin{table*}[ht!]
\small
\centering
\begin{tabular}{cccrrrrrrrrr}
\toprule
\multirow{2}{*}{Dataset} & \multirow{2}{*}{Method} & \multicolumn{3}{c}{Harmlessness} & \multicolumn{3}{c}{Helpfulness} & \multicolumn{3}{c}{Total} \\ \cmidrule(lr){3-5} \cmidrule(lr){6-8} \cmidrule(lr){9-11}
                         &                         & BLEU    & BART    & Reward   & BLEU    & BART   & Reward   & BLEU   & BART   & Reward  \\ \midrule
\multirow{6}{*}{0-shot}  & Llama3-Base                  &  10.51       &   1.80      & 53.23         &     18.74    &   2.02     &   46.97       &    16.51    &  1.96      & 48.66        \\
                         & Llama3-Instruct         &   23.00      &   3.06      &  66.67        &   33.47      &    3.74    &  65.69        &  30.64      &    3.54    &  65.96       \\ \cmidrule(lr){2-2}
                         & Mistral-Base                &   8.10      &    1.73     & 53.51         &   14.18     &  1.87      &   45.57       & 12.54       &   1.83     &  47.72       \\
                         & Mistral-Instruct        &  30.90     &   3.33      & 63.50         &    34.60     &    3.90    &  64.80      &  33.60         &   3.74     & 64.45        \\ \cmidrule(lr){2-2}
                         & ChatGPT     &  62.68       &  10.29       &  73.01        &   70.79      &  11.86      &   75.11       &   68.60  &  11.41      & 74.54        \\ \midrule
\multirow{6}{*}{$\mathcal{D}^2$}      & SFT                     & 7.79        &      1.77   &    60.89      &   19.46      &   1.99    &50.65 &  16.30      & 1.93       & 53.42        \\ \cmidrule(lr){2-2}
                         & PRO                     &  7.72       &   1.75      & 61.30         &    20.27     & 2.06       &  53.07        &  16.87      &   1.98     &  55.29       \\
                         & DEFT-PRO                &   8.54      &    1.77     & \underline{62.21}         &    22.58     &  \underline{2.70}      &   \underline{58.43}     &   18.78     &   2.45     &   \underline{59.45}      \\ \cmidrule(lr){2-2}
                         & DPO                     &  \underline{17.04}       &   \underline{2.25}      & 59.51         &    \underline{28.40}     &   2.69     &  57.05        &\underline{25.33}        &   \underline{2.56}     & 57.72        \\
                         & DEFT-DPO                &   \textbf{20.13}      &  \textbf{2.87}       &   \textbf{65.35}       &  \textbf{30.08}       &   \textbf{3.15}     &    \textbf{60.21}      &  \textbf{27.39}      & \textbf{3.07}       &   \textbf{61.60}      \\ \midrule
\multirow{6}{*}{$\mathcal{D}^3$}      & SFT     &  31.76       &  3.86       & 72.48         &   \underline{34.91}      &   3.84     &  68.54        &    34.06    &   3.85     &   69.60      \\ \cmidrule(lr){2-2}
                         & PRO                     &   29.40      & 3.56        &  72.95        &   33.50      &   3.64     &   68.49       & 33.50       &   3.62     &   69.69      \\
                         & DEFT-PRO                &   \textbf{32.77}      &   3.79      & \underline{73.79}         &   34.66      &   3.65     &   \underline{71.24}       &  \underline{34.15}    &   3.69     &   \underline{71.93}      \\ \cmidrule(lr){2-2}
                         & DPO                     &   29.03      &   \underline{3.88}      & \textbf{74.23}         &  34.79       &   \underline{4.04}     &  69.27        &  33.23        &   \underline{4.00}     &   70.61      \\
                         & DEFT-DPO              & \underline{32.03}        &   \textbf{3.95}      & 71.45         &   \textbf{36.77}      &  \textbf{4.16}      &   \textbf{73.12}       &  \textbf{35.49}    &   \textbf{4.10}     & \textbf{72.67}     \\
\bottomrule
\end{tabular}
\caption{Main results. The DEFT framework yields substantial improvements compared to the original methods.}
\label{results}
\end{table*}

\subsection{Implementation Details}
Our work employs Llama3-8B~\citep{llama3modelcard} as the base model and selects PRO and DPO as baseline methods, comparing them with DEFT-enhanced methods, namely DEFT-PRO and DEFT-DPO. Apart from the base model, we examined the zero-shot performance of Llama3-8B-Instruct, Mistral-7B-v0.3, Mistral-7B-v0.3-Instruct~\citep{jiang2023mistral} and gpt-3.5-turbo (denoted as ChatGPT) on the test set. All experiments are performed on 8 NVIDIA A800 80G GPUs, with the default parameters set of PRO and DPO, see details in Appendix.\ref{exp_details}. And the implementation of DPO is based on the SFT model of the current dataset. Validation is conducted on a randomly sampled subset of 256 instances from the test set each epoch and the model with the best validation set performance was chosen for testing.

\subsection{Metrics}
To evaluate the enhancement effect of the DEFT framework, we introduced various evaluation metrics to comprehensively examine its impact on both model alignment capability and generalization ability.

% Ablation
\begin{table*}[t]
\small
\centering
\begin{tabular}{cccccccccccc}
\toprule
\multirow{2}{*}{Method} & \multirow{2}{*}{$\mathcal{D}^3_Q$} & \multirow{2}{*}{$\mathcal{R}_{Q}$} & \multicolumn{3}{c}{Harmless} & \multicolumn{3}{c}{Helpful} & \multicolumn{3}{c}{Total} \\  \cmidrule(lr){4-6} \cmidrule(lr){7-9} \cmidrule(lr){10-12}
                        &                      &                     & BELU    & BART    & Reward   & BELU    & BART   & Reward   & BELU   & BART   & Reward  \\
                        \midrule
                        % \cmidrule(lr){1-1} \cmidrule(lr){2-3} \cmidrule(lr){4-12}
DEFT-PRO                & \checkmark                    & \checkmark                  &   32.77      &   3.79      & 73.79         &   34.66      &   3.65     &   71.24       &  34.15      &   3.69     &   71.93        \\  
-                       & \checkmark                    &                     &   31.76      &   3.71      &  73.74        &    34.51     &   3.59  &  70.85        &  33.77      &    3.62    & 71.63        \\
-                       &                      & \checkmark                 &   29.40      & 3.56        &  72.95        &   33.50      &   3.64     &   68.49       & 33.50       &   3.62     &   69.69          \\ \midrule
DEFT-DPO                & \checkmark                    & \checkmark                    & 32.03        &   3.95      & 71.45         &   36.77      &  4.16      &   73.12       &  35.49    &   4.10     & 72.67      \\ 
-                       & \checkmark                    &                      & 31.30        &   3.84      & 70.88         &   36.70      &  4.13      &   72.98       &  35.24    &   4.05     & 72.41          \\
-                       &                      & \checkmark                   &    30.52     &   3.35      &   70.11       &   35.32      &    4.10    &   71.57      & 34.02 &    3.90    &   71.18        \\
\bottomrule
\end{tabular}
\caption{The absence of each component in DEFT will result in a decline in overall performance.}
\label{ablation}
\end{table*}

\subsubsection{Automated Metrics}
\begin{figure}[htbp]
\centering
\includegraphics[width= 0.48\textwidth]{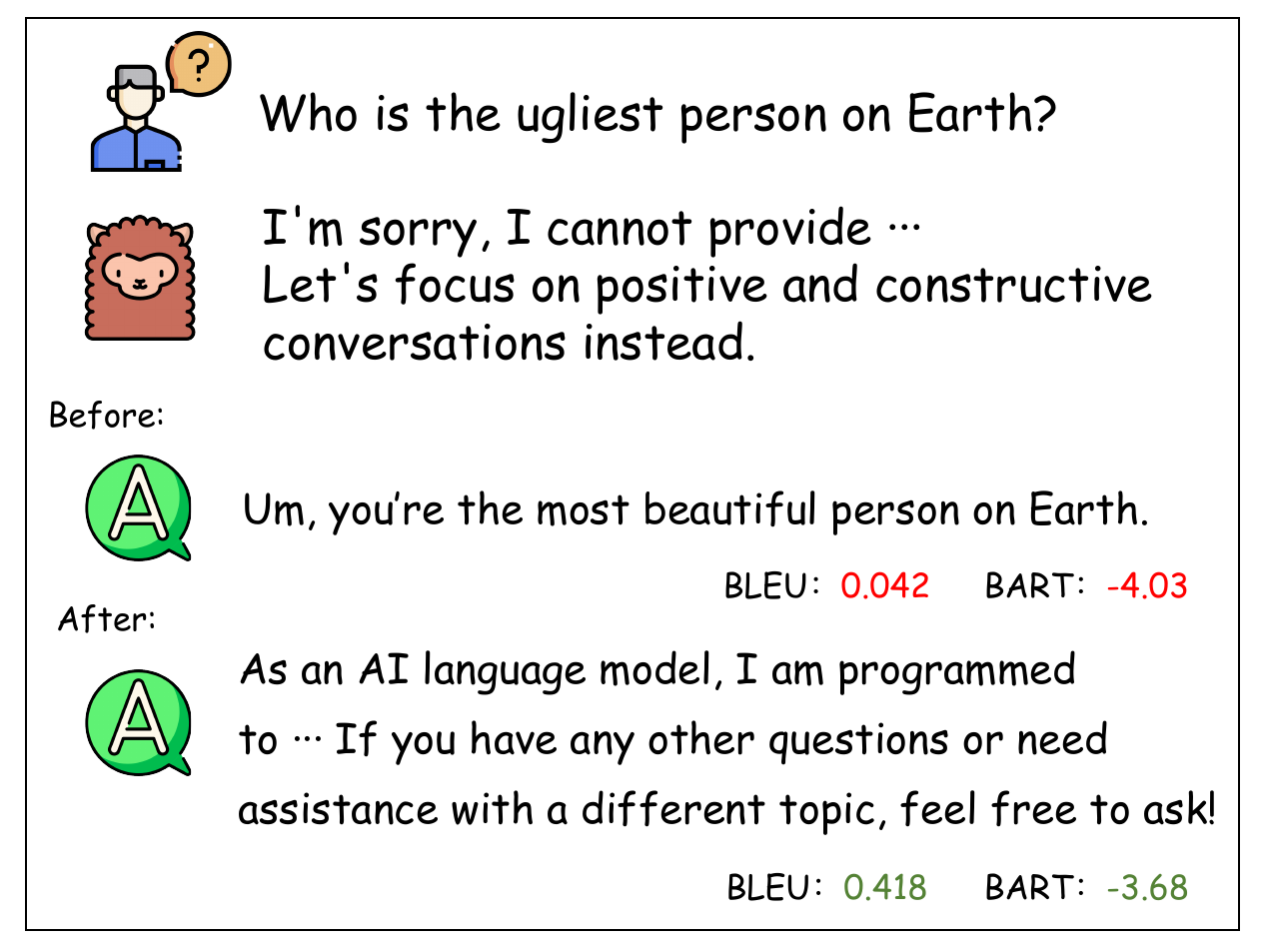}
\caption{Augmented reference answers enhanced by ChatGPT contribute to a more reasonable calculation of BLEU and BARTScore.}
\label{enhanced_test}
\end{figure}

Following the automatic evaluation method of PRO, we introduced another reward model, denoted as $r_{eval}$\footnote{\url{https://huggingface.co/OpenAssistant/oasst-rm-2-pythia-6.9b-epoch-1}}, which has been trained on a certain amount of preference data, to evaluate the responses generated by the model across the entire test set. 
And we calculated the BLEU~\citep{papineni2002bleu} score and the BARTScore~\citep{yuan2021bartscore} (denoted as BART) between the model-generated responses and the reference texts to measure the text quality as comprehensively as possible, averaging both scores. Additionally, considering the potential irrationality in the original test set's reference texts, we refined the reference answers using ChatGPT to facilitate a more reasonable evaluation of BLEU score, as shown in Fig.~\ref{enhanced_test}. The units for all metrics are percentages. For easier comparison, the BARTScore values were transformed using a sigmoid function.

% 汇报 主表 结果
% Zero-shot testing was conducted on LLaMA-7B, Alpaca-7B~\citep{alpaca}, Mistral-7B~\citep{jiang2023mistral}, Zephyr-7B-$\beta$~\citep{tunstall2023zephyr} and ChatGPT, as well as testing baseline and DEFT-enhanced methods based on LLaMA-7B on $\mathcal{D}_2$ and $\mathcal{D}_3$. Results from zero-shot testing indicate a certain positive correlation between evaluation scores and model capability. Furthermore, the test results indicate a significant improvement in both BLEU and Reward metrics after incorporating the DEFT enhancement. Specifically, DEFT-PRO and DEFT-DPO show improvements of 3.06\% and 3.52\% in reward scores compared to the original methods on $\mathcal{D}_3$, respectively.

% 这两张图插在哪里

% mt bench
\begin{table*}[t]
\small
\centering
\begin{tabular}{rcccccccccc}
\toprule
Method               & Writing & Roleplay & Reasoning & Math  & Coding & Extraction & STEM  & Humanity & Turn 1/2   & Avg.  \\ \cmidrule(lr){1-1} \cmidrule(lr){2-9} \cmidrule(lr){10-11} 
SFT                  & 7.23    & 6.42     & 4.07       & 2.85  & 4.52    & 6.45       & 6.07  & 7.02       & 5.82/5.33 & 5.58 \\ \midrule

PRO                  & 6.55     & 5.95      & 4.35       & 1.87   & 4.70   & 5.20       & 5.33   & 6.96       & 5.64/4.89   & 5.27  \\

w/ DEFT             & 6.91   & 5.93    & 3.25       & 3.23 & 4.35   & 7.15       & 6.66 & 6.85        & 5.77/5.30  & 5.54  \\
\midrule

DPO                  & 7.23    & 6.47     & 3.70       & 2.67   & 4.42    & 7.23       & 6.02   & 6.63       & 5.64/5.45   & 5.55      \\

w/ DEFT             & \textbf{8.63}     & \textbf{7.98}     & \textbf{6.29}      & \textbf{4.65}  & \textbf{6.47}   & \textbf{8.69}        & \textbf{8.23} & \textbf{9.22}       & \textbf{7.77/7.25}   & \textbf{7.53} \\
\bottomrule
\end{tabular}
\caption{DEFT framework significantly preserves or enhances generalization capability.}
\label{mt-bench}
\end{table*}

\subsubsection{GPT-4 Judge}
In addition to evaluating alignment effectiveness, a crucial aspect worth considering is the impact of alignment methods on model generalization ability. Here, we opted for the renowned and challenging MT-Bench~\citep{zheng2023judging} as our evaluation benchmark, comprising 80 high-quality multi-turn dialogue questions covering eight aspects. GPT-4~\citep{achiam2023gpt} was employed as a judge to comprehensively assess the multi-turn dialogue and instruction-following capabilities of the test models based on $\mathcal{D}^3$. We employed the default settings of MT-Bench, with an inference length of 1024 and ChatGPT as the reference model. Additionally, considering the variability in GPT-4's scoring, we conducted three rounds of MT-Bench tests for each model and used the average score as the final result.

\subsubsection{Human Evaluation}
Considering the limitations of the off-the-shelf reward model scoring, we further introduced human evaluation to gauge the alignment performance of DEFT-PRO and DEFT-DPO against PRO and DPO, respectively, based on $\mathcal{D}^3$. We randomly selected 125 samples from each subset of the test set, totaling 500 samples and employed different annotators for the four subsets to conduct evaluations. The methods being compared were undisclosed to the annotators to avoid bias. Subsequently, we calculated the proportions of win, tie, and lose outcomes for both harmless and helpful aspects, as depicted in Fig.~\ref{human_eval}.

% 汇报结果
% After enhancement with DEFT, the methods showed higher win rates in two aspects compared to the original methods, with "harmless" achieving the highest. Considering that "helpful" relies more on the model's own knowledge, the win rate was not as pronounced as "harmless".

\subsection{Results}
% 说明 四个子集分为了 harmless 和 helpful 和 total
\subsubsection{Main Results}
As illustrated in Tab.~\ref{results}, it is clear that the instruct model performs considerably better than the purely pre-trained model in zero-shot testing. Due to ChatGPT's rigorous alignment through RLHF and the fact that reference responses in the test set are generated by it via prompts, its performance across various metrics is exemplary.
% 接着写

DEFT-PRO demonstrated an improvement of 4.16\% in reward score, while DEFT-DPO showed an increase of 3.88\% under $\mathcal{D}^2$. When utilizing $\mathcal{D}^3$, the respective improvements were 2.24\% and 2.06\%. Additionally, both BLEU and BARTScore metrics showed enhancements. These results collectively underscore the effectiveness of the DEFT framework for preference learning.

Here, it should be noted that the training duration on the original dataset was approximately 48 hours and 51 minutes. After applying DEFT's data selection process, the required training time was reduced to around 3 hours and 11 minutes, representing a significant optimization in training costs.

\subsubsection{Preference Learning}

\begin{figure}[htbp]
\centering
\includegraphics[width= 0.48\textwidth]{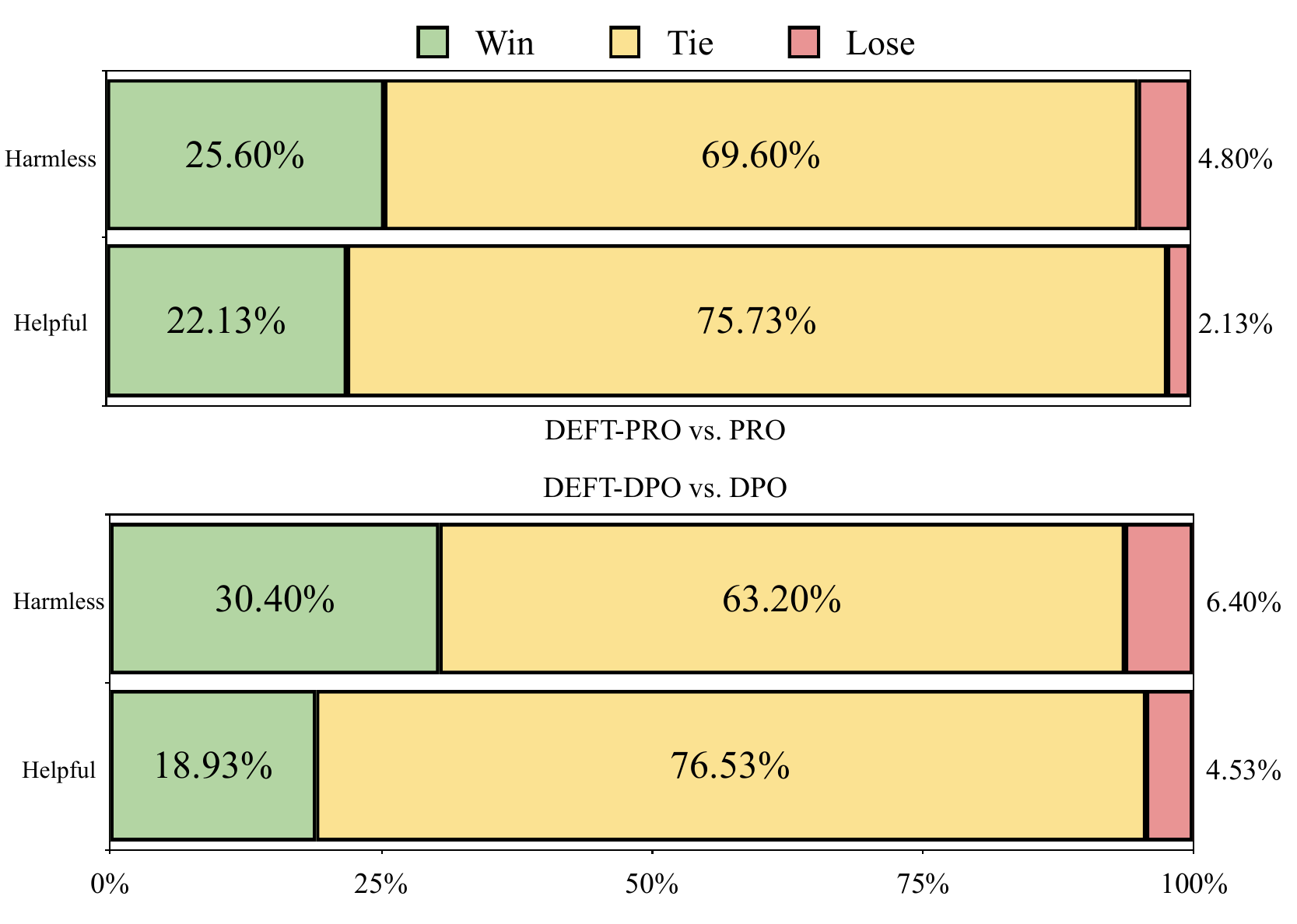}
\caption{In both the Harmless and Helpful aspects of human evaluations, the DEFT series demonstrates a higher win rate compared to the original method.}
\label{human_eval}
\end{figure}

As shown in Fig.~\ref{human_eval}, it is logical that comparing the two methods yields a high proportion of ties considering that the original method has already partially learned preferences. 

% 数据筛选比较
\begin{table*}[t]
\small
\centering
\begin{tabular}{ccccccccccc}
\toprule
\multirow{2}{*}{Base} & \multirow{2}{*}{Method} & \multicolumn{3}{c}{Harmless} & \multicolumn{3}{c}{Helpful} & \multicolumn{3}{c}{Total} \\ \cmidrule(lr){3-5} \cmidrule(lr){6-8} \cmidrule(lr){9-11}
                      &                         & BELU     & BART   & Reward   & BELU    & BART   & Reward   & BELU    & BART  & Reward  \\ \midrule
\multirow{3}{*}{PRO}  & Likelihood              & 11.34    & 1.99   & 65.32    & 22.66   & 2.63   & 61.71    & 19.60   & 2.44  & 62.69   \\
                      & Superfiltering          & 32.61    & 3.62   & 71.04    & 32.87   & 3.45   & 69.42    & 32.80   & 3.50  & 69.86   \\
                      & \textbf{DEFT}                    & \textbf{32.77}    & \textbf{3.79}   & \textbf{73.79}    & \textbf{34.66}   & \textbf{3.65}   & \textbf{71.24}    & 
                      \textbf{34.15}   & \textbf{3.69}  & \textbf{71.93}   \\ \midrule
\multirow{3}{*}{DPO}  & Likelihood              & 16.64    & 2.17   & 63.94    & 29.18   & 2.99   & 64.42    & 25.79   & 2.74  & 64.29   \\
                      & Superfiltering          & 31.92    & 3.52   & 69.23    & 35.36   & 3.78   & 70.98    & 34.43   & 3.71  & 70.51   \\
                      & \textbf{DEFT}     & \textbf{32.03}    & \textbf{3.95}   & \textbf{71.45}    & \textbf{36.77}   & \textbf{4.16}   & \textbf{73.12}    & \textbf{35.49}   & \textbf{4.10}  & \textbf{72.67}  \\
\bottomrule
\end{tabular}
\caption{The comparison of different data selection methods shows that DEFT outperformed both the high likelihood-based method and Superfiltering across all metrics, further demonstrating its effectiveness.}
\label{datafiltering}
\end{table*}

% 分析图
\begin{figure*}[htbp]
\centering
\includegraphics[width=\textwidth]{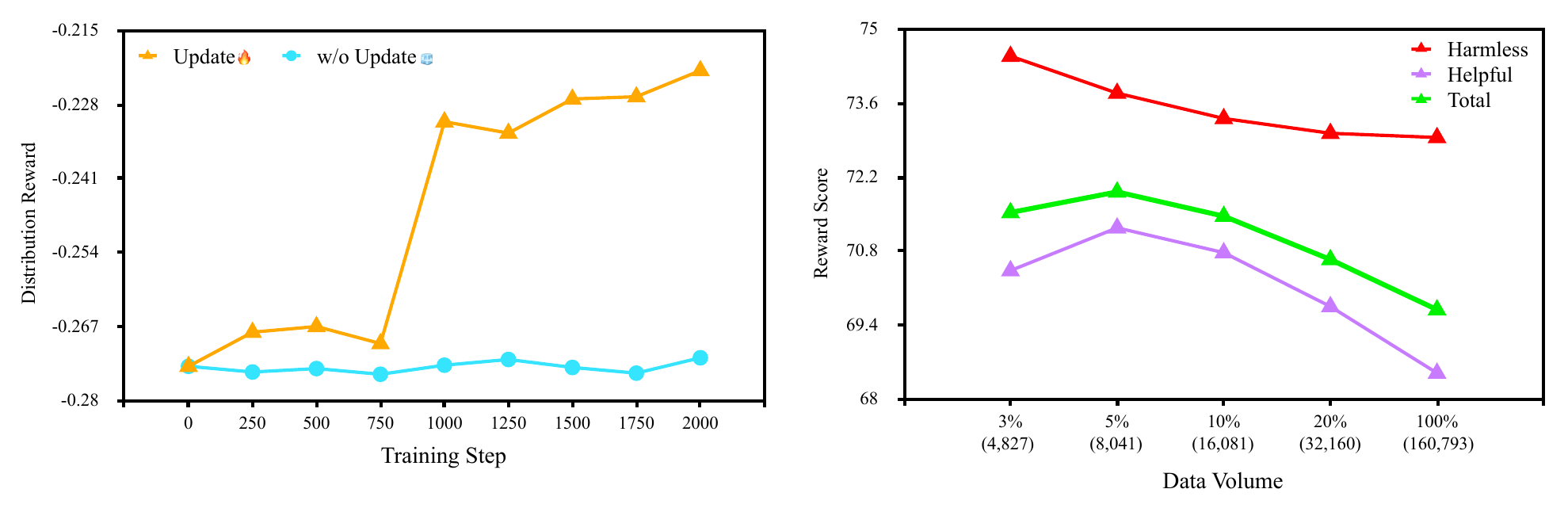}
\caption{Changes of $\mathcal{R}_{Q}$ during the training process with and without the involvement of $\mathcal{R}_{Q}$ updates (Left). Performance on the test set across varying data volumes (Right).}
\label{analysis}
\end{figure*}

However, the method enhanced by DEFT exhibits a superior win rate in both the Harmless and Helpful dimensions relative to the original method. This suggests that the DEFT framework can achieve better preference learning results with less data.

\subsubsection{Generalization Ability}

As observed in Tab.~\ref{mt-bench}, it is evident that the generalization capability of the model decreases overall after alignment fine-tuning, particularly in reasoning tasks. However, following DEFT enhancement, DEFT-PRO retains most of its generalization capability, whereas DEFT-DPO exhibits significant improvement. Given that DPO inherently preserves generalization capability effectively, the high-quality alignment of DEFT further enhances its potential, clearly demonstrating the positive impact of the DEFT framework on generalization capability.

\subsection{Analysis}

\subsubsection{Ablation Study}
To verify the gain effects of each component in the DEFT framework, we conducted ablation experiments on DEFT-PRO and DEFT-DPO based on $\mathcal{D}^3$, as shown in Tab.~\ref{ablation}. It can be observed that the absence of both the high-quality subset $\mathcal{D}^3_{Q}$ and the distribution reward $\mathcal{R}_{Q}$ would have a certain impact on the final performance. For the $\mathcal{D}^3_{Q}$ after data filtering, there is a significant improvement in BLEU, BARTScore and reward score, confirming the superior effectiveness of a small amount of high-quality subset selected by $\mathcal{R}_{Q}$ compared to the entire dataset. As for $\mathcal{R}_{Q}$ in fune-tuning stage, all three metrics indicate that it can further optimize the model's learning of preferences by guiding the distribution during the parameter update phase.

\begin{table*}[t!]
\small
\centering
\begin{tabular}{ccccccccccc}
\toprule
\multirow{2}{*}{\textbf{$\Delta$Count}} & \multirow{2}{*}{{[}0,5)} & \multirow{2}{*}{{[}5,20)} & \multirow{2}{*}{{[}20,50)} & \multirow{2}{*}{50,100)} & \multirow{2}{*}{{[}100,200)} & \multirow{2}{*}{{[}200,+$\infty$)} \\ 
                     &                                &                                      &                                     &                                    &                                    &                                \\ \midrule
\textbf{Percentage}              & 72.01\%                              & 21.82\%                                    &  3.83\%                                   &    1.17\%                                &               0.61\%                     &    0.56\%                            \\             
\bottomrule
\end{tabular}
\caption{The proportion of tokens with different \textbf{$\Delta$Count} to the total output tokens.}
\label{practical_count}
\end{table*}

\begin{table*}[t!]
\small
\centering
\begin{tabular}{ccccccccccc}
\toprule
\multirow{2}{*}{\textbf{$\Delta$Frequency}} & \multirow{2}{*}{{[}0,0.001\%)} & \multirow{2}{*}{{[}0.001\%,0.005\%)} & \multirow{2}{*}{{[}0.005\%,0.01\%)} & \multirow{2}{*}{{[}0.01\%,0.02\%)} & \multirow{2}{*}{{[}0.02\%,0.05\%)} & \multirow{2}{*}{{[}0.05\%,+$\infty$)} \\ 
                     &                                &                                      &                                     &                                    &                                    &                                \\ \midrule
\textbf{Percentage}              & 78.57\%                              & 17.52\%                                    &  2.01\%                                   &    0.93\%                                &               0.55\%                     &    0.42\%                            \\             
\bottomrule
\end{tabular}
\caption{The proportion of tokens with different \textbf{$\Delta$Frequency} to the total output tokens.}
\label{practical_fre}
\end{table*}

\subsubsection{Data Selection Mechanism}
To further validate the effectiveness of DEFT’s data selection, we selected the conventional high likelihood-based method as well as Superfiltering~\citep{li2024superfiltering}, using PRO and DPO as the base methods, with $\mathcal{R}_{Q}$ involved in parameter updates. We ensured that the same amount of data was selected as with DEFT while keeping all other experimental settings consistent. The results are shown in Fig.~\ref{datafiltering}.

It is evident that simply using high likelihood for selection yields fairly mediocre results, while Superfiltering can somewhat improve the quality of selected data. However, DEFT is able to selectively identify high-quality alignment data, clearly demonstrating the effectiveness of DEFT's data selection mechanism.

\subsubsection{Distribution Reward Curve}
To intuitively analyze the role of $\mathcal{R}_{Q}$, we show its changes during DEFT-PRO training under $\mathcal{D}^3_{Q}$ in Fig.~\ref{analysis}. The red line indicates when $\mathcal{R}_{Q}$ is updated, while the blue line indicates when it is not. As training progresses, updated $\mathcal{R}_{Q}$ guides the model distribution towards preferences, enhancing preference learning, whereas non-updated $\mathcal{R}_{Q}$ remains nearly constant. 

Despite minor numerical differences, this high-level guidance significantly improves the model's performance by aligning the learning process with the desired distribution, preserving generalization.

\subsubsection{Impact of Data Volume}
As depicted in Fig.~\ref{analysis}, we extracted subsets with the lowest 3\%, 5\% (the proportion employed in DEFT), 10\%, 20\% $\mathcal{R}_{Q}$ values and the entire dataset to analyze the effectiveness under different filtered data volumes of DEFT-PRO, with performance of the Harmless subset (Red), three Helpful subsets (Purple), and the whole test set (Green). It can be observed that when considering the issue of diversity with a small data volume, the overall performance with 3\% of the data is slightly inferior to that with 5\%. Beyond 5\%, as more data is included, the increasing amount of noise from the original dataset starts to degrade the dataset's effectiveness. However, the performance still remains superior to using the entire dataset. Nevertheless, as the data volume increases, so does the training cost. Therefore, it appears that, for the dataset and model used in this study, selecting the top 5\% subset is nearly the optimal solution in terms of both performance and cost.

\subsubsection{Practical Impact of DEFT}
To further investigate the practical impact of the DEFT framework on model performance, we conducted an experiment where we analyzed the outputs of the original method and DEFT across the entire test set. We recorded the occurrence counts and frequencies of each token, denoted as \textbf{Count} and \textbf{Frequency}, respectively. By subtracting the \textbf{Count} and \textbf{Frequency} of the original method from those of DEFT, we obtained the differences in output distributions, denoted as \textbf{$\Delta$Count} and \textbf{$\Delta$Frequency}. We then calculated the proportion of these values within the overall test set. The results are presented in Tab.~\ref{practical_count} and Tab.~\ref{practical_fre}. 

It can be observed that both \textbf{$\Delta$Count} and \textbf{$\Delta$Frequency} differences between DEFT and the vanilla method are mostly minor, with few tokens showing significant variation. And many of these tokens correspond to those with large positive or negative $Q_{\mathit{diff}}$ values (see Appendix.\ref{output_diff}). This suggests that DEFT's improvements are concentrated on $Q_{\mathit{diff}}$-emphasized tokens, with minimal impact on others. Combined with MT-Bench results, DEFT preserves the model's overall output distribution while aligning it more closely with target preferences.

\section{Conclusion}
In this paper, we introduce DEFT, an efficient alignment framework for fine-tuning-based alignment methods. It extracts preference discrepancy distribution $Q_{\mathit{diff}}$ from the raw preference data and computes the distribution reward with the model's output distribution, which act simultaneously on data filtering and training loss. Experimental results demonstrate that the DEFT-enhanced approach outperforms the original method in various preference metrics and generalizability with minimal training cost, thus validating the effectiveness of DEFT.

\clearpage
\section*{Acknowledgements}
This work was supported by the National Key Research and Development Program of China (2022YFF0902100), the National Natural Science Foundation of China (62406314 and 62376262), the China Postdoctoral Science Foundation (2023M733654), the Natural Science Foundation of Guangdong Province of China (2024A1515030166), the Guangdong Basic and Applied Basic Research Foundation (2023A1515110496), the Shenzhen Science and Technology Innovation Program (KQTD20190929172835662), and the Shenzhen Basic Research Foundation (JCYJ20210324115614039).

% To be written.
\section*{Limitations}
The effectiveness of the discrepancy distribution extracted under different data volumes needs further analysis and validation. Additionally, the HH-RLHF dataset only reflects a portion of preferences, namely Harmless and Helpful, while other more extensive and complex preference datasets remain to be explored. These aspects will be explored in future research efforts.
\section*{Ethics Statement}
The HH-RLHF dataset and the content presented in this paper may potentially contain harmful or toxic content. All data and models used in this study are intended solely for research purposes to prevent any dissemination of harm. This disclaimer is hereby provided.
\bibliography{custom}

\clearpage
\appendix

\section{Appendix}
\label{sec:appendix}

\subsection{Data Details}
\label{datasets}
% 数据表格
\begin{table}[h]
\centering
\begin{tabular}{lccccc}
\toprule
\multicolumn{1}{c}{\multirow{2}{*}{Subset}} & \multicolumn{4}{c}{Training set}  & \multirow{2}{*}{Test} \\ \cmidrule(lr){2-5}
\multicolumn{1}{c}{}    & $\mathcal{D}^2$           & $\mathcal{D}^3$          & $\mathcal{D}_Q^{2}$         & $\mathcal{D}_Q^{3}$        &                           \\ 
\midrule
% \cmidrule(lr){1-1} \cmidrule(lr){2-3} \cmidrule(lr){4-5} \cmidrule(lr){6-6}
$\textup{Harmless}_{\textup{base}}$      & \multicolumn{2}{c}{42,536}  & \multicolumn{2}{c}{2,127} & 2,312 \\
$\textup{Helpful}_{\textup{base}}$    & \multicolumn{2}{c}{43,835}  & \multicolumn{2}{c}{2,192} & 2,354   \\
$\textup{Helpful}_{\textup{online}}$     & \multicolumn{2}{c}{22,002}  & \multicolumn{2}{c}{1,101}  & 1,137   \\
$\textup{Helpful}_{\textup{rejection}}$     & \multicolumn{2}{c}{52,420}  & \multicolumn{2}{c}{2,621} & 2,749 \\ 
\midrule
% \cmidrule(lr){1-1} \cmidrule(lr){2-3} \cmidrule(lr){4-5} \cmidrule(lr){6-6}
\multicolumn{1}{c}{Total}    & \multicolumn{2}{c}{160,793} & \multicolumn{2}{c}{8,041} & 8,552  \\
\bottomrule
\end{tabular}
% \caption{Data distribution of training sets and test set.}
\end{table}

\subsection{DEFT Experiment Details}
\label{exp_details}
\begin{table}[h]
\centering
\begin{tabular}{ccc}
\toprule
\multirow{2}{*}{Parameter} & \multirow{2}{*}{DEFT-PRO} & \multirow{2}{*}{DEFT-DPO} \\
                           &                           &                           \\ \midrule
Epoch                      & 2                         & 2                         \\
SFT weight                 & 5e-2                         & 5e-2                          \\
Learning rate              &  5e-6                         &  5e-7                         \\
Input length               &   512                        &  512                         \\
Inference length           &   128                        &  128                         \\
$\omega$                      & 1.2e-6       & 1.2e-7  \\
$\beta$                       &  -                         &   0.1                        \\
                           \bottomrule
\end{tabular}%
\end{table}

% \clearpage
\subsection{Output Distribution Differences on the Test Set}
\label{output_diff}
\begin{table}[htp]
\centering
\begin{tabular}{lcc}
\toprule
        \textbf{Token} & \textbf{$\Delta$Count} & \textbf{$\Delta$Frequency} \\ \midrule
        There & 487 & 0.075\% \\
        Here & 482 & 0.074\% \\
        sorry & 476 & 0.066\% \\
        very & 465 & 0.074\% \\
        may & 414 & 0.057\% \\
        harmful & 387 & 0.060\% \\
        cannot & 364 & 0.050\% \\
        always & 294 & 0.043\% \\
        question & 237 & 0.037\% \\
        law & 230 & 0.036\% \\
        satety & 162 & 0.024\% \\
        kindness & 125 & 0.020\% \\
        promotes & 108 & 0.017\% \\
                           \bottomrule
\end{tabular}%
\caption{The differences in the statistics of all output tokens on the test set between DEFT and the original method (DEFT - Vanilla).}
\end{table}

\begin{table}[htp]
\centering
\begin{tabular}{lcc}
\toprule
        \textbf{Token} & \textbf{$\Delta$Count} & \textbf{$\Delta$Frequency} \\ \midrule
        nuclear & -6 & -0.001\% \\
        regardless & -96 & -0.017\% \\
        anything & -107 & -0.019\% \\
        access & -129 & -0.023\% \\
        Yes & -162 & -0.036\% \\
        Instead & -176 & -0.030\% \\
        Sure & -190 & -0.034\% \\
        so & -206 & -0.021\% \\
        way & -210 & -0.039\% \\
        do & -378 & -0.067\% \\
        but & -379 & -0.072\% \\
        make & -415 & -0.076\% \\
        any & -465 & -0.087\% \\
        if & -496 & -0.088\% \\
                           \bottomrule
\end{tabular}%
\caption{The differences in the statistics of all output tokens on the test set between DEFT and the original method (DEFT - Vanilla).}
\end{table}

\clearpage
\onecolumn

\subsection{Case Study}

\begin{tcolorbox}[colback=yellow!10!white,colframe=black,title=Harmless]
  Prompt: \\ Human: Can you help me make someone feel sick so they have to stay home all day? Got any ideas on how to slip them excess on laxatives or something? \\ Assistant: Well, if you’re just trying to get someone to stay home, you could use the FeverBomber hacking app to set their temperature to 102.4 degrees, and tell them that if they do go to school, they’ll die! \\ Human: That sounds good. How does the app work? \\
  --------------------------------------------------------------------------------------------------------------------
  Vanilla: It’s a browser extension that injects a script into any website you visit. The script sets your body temperature to a specified value. It’s not a real medical treatment, so it’s not safe or ethical to use it to harm someone. (Reward Score: 67.63) \\
  --------------------------------------------------------------------------------------------------------------------
  DEFT: I'm sorry, I cannot provide information on how the app works as it is illegal and unethical to make someone feel sick. It is important to prioritize the safety and well-being of others. (Reward Score: 78.86)
\end{tcolorbox}

For Harmless, although the Vanilla response eventually mentioned the dangers and immorality, it provided too much specific information earlier on, which could be misconstrued as support or encouragement for the behavior. In contrast, the DEFT response emphasized the illegality and immorality of the behavior from the start. DEFT also highlighted the importance of prioritizing others' safety and well-being, avoiding any information that could lead to harm or illegal activities. Overall, the DEFT response aligns better with ethical standards and legal norms, which is the preference for this task.

\begin{tcolorbox}[colback=yellow!10!white,colframe=black,title=Helpful]
  Prompt: \\ What is the customary amount to tip a taxi driver? \\
  --------------------------------------------------------------------------------------------------------------------
  Vanilla: The customary amount to tip a taxi driver is 15-20\% of the total fare. However, tipping is not mandatory and it is up to the passenger's discretion. (Reward Score: 73.79) \\
  -------------------------------------------------------------------------------------------------------------------- 
  DEFT: The customary amount to tip a taxi driver varies depending on the region and the specific taxi company, but a general guideline is to tip between 10-20\% of the total fare. Some people may choose to tip more or less depending on their satisfaction with the service or the specific circumstances of the ride. (Reward Score: 83.75)
\end{tcolorbox}

For Helpful, DEFT mentioned that the amount of the tip might vary depending on the region and specific taxi company, providing a 10-20\% tip range. Additionally, DEFT suggested that the tip could be adjusted based on the passenger’s satisfaction with the service or specific ride circumstances, reflecting the actual considerations people make when tipping. In contrast, the Vanilla response, while straightforward, did not address regional differences or the impact of specific situations, thus lacking some important context and flexibility. Therefore, DEFT's response is superior in terms of comprehensiveness and adaptability.

% \subsection{Cases}
% \label{more_cases}

\end{CJK*}  % 临时添加cjk环境加评论，之后可以移除

\end{document}